\documentclass{article}

    \PassOptionsToPackage{numbers, compress}{natbib}
\usepackage[preprint]{neurips_2025}

\usepackage[utf8]{inputenc} 
\usepackage[T1]{fontenc}    
\usepackage{hyperref}       
\usepackage{url}            
\usepackage{booktabs}       
\usepackage{amsfonts}       
\usepackage{nicefrac}       
\usepackage{microtype}      
\usepackage{xcolor}         

\usepackage{multirow}
\usepackage{multicol}
\usepackage{booktabs}
\usepackage{subfigure}
\usepackage[subfigure]{tocloft}
\usepackage{subcaption}
\usepackage{hhline}
\usepackage[ruled, vlined]{algorithm2e}
\usepackage{algpseudocode}
\usepackage{adjustbox}
\usepackage{graphicx}
\usepackage{balance}
\usepackage{caption} 
\usepackage[most]{tcolorbox}
\usepackage{color, colortbl}
\usepackage{amsthm}
\usepackage{anyfontsize}
\usepackage{siunitx}
\usepackage{wrapfig}
\usepackage[para,online,flushleft]{threeparttable}
\usepackage{tikz}
    \usetikzlibrary{fadings}
    \usetikzlibrary{patterns}
    \usetikzlibrary{shadows.blur}
    \usetikzlibrary{shapes}
\usepackage{pdflscape}
\usepackage{tabularx}
\usepackage[toc,page]{appendix}
\usepackage[withpage, smaller, nohyperlinks]{acronym} 
\usepackage{enumitem,amssymb}
\usepackage{pifont}
\usepackage[para,online,flushleft]{threeparttable}
\usepackage[final]{pdfpages}
\usepackage[capitalize,noabbrev]{cleveref}
\usepackage{threeparttable}
\usepackage{tcolorbox}
\usepackage{colortbl}

\usepackage{tikz}
\usepackage{tikz-qtree}
\usepackage{pgfplots}

\definecolor{jkublue}{RGB}{0, 100, 138}

\hypersetup{
   colorlinks=true,
   linkcolor=jkublue,
   citecolor=jkublue,
   urlcolor=black
}

\renewcommand{\cite}{\citep}

\definecolor{RedOrange}{RGB}{255,69,0}

\definecolor{lightblue}{RGB}{32, 194, 217}
\definecolor{cell_red}{RGB}{217, 92, 76}
\definecolor{cell_blue}{RGB}{0, 132, 187}
\definecolor{cell_green}{RGB}{91, 167, 85} %
\definecolor{cell_yellow}{RGB}{241, 188, 63}
\definecolor{cell_cyan}{RGB}{79,176,191}
\definecolor{cell_grey}{RGB}{125,130,140}
\definecolor{cell_lightgreen}{RGB}{191,206,82}
\definecolor{cell_violett}{RGB}{174,97,157}

\usepackage{cleveref}
\crefname{section}{§}{§§}
\Crefname{section}{§}{§§}

\title{From Static Structures to Ensembles: Studying and Harnessing Protein Structure Tokenization}

\author{%
  Zijing Liu\thanks{\texttt{liuzijing@idea.edu.cn}} , Bin Feng, He Cao, Yu Li  \\
  International Digital Economy Academy\\
  Shenzhen, China \\
}

\begin{document}

\maketitle

\begin{abstract}
Protein structure tokenization converts 3D structures into discrete or vectorized representations, enabling the integration of structural and sequence data. Despite many recent works on structure tokenization, the properties of the underlying discrete representations are not well understood. In this work, we first demonstrate that the successful utilization of structural tokens in a language model for structure prediction depends on using rich, pre-trained sequence embeddings to bridge the semantic gap between the sequence and structural ``language''. The analysis of the structural vocabulary itself then reveals significant semantic redundancy, where multiple distinct tokens correspond to nearly identical local geometries, acting as ``structural synonyms''. This redundancy, rather than being a flaw, can be exploited with a simple ``synonym swap'' strategy to generate diverse conformational ensembles by perturbing a predicted structure with its structural synonyms. This computationally lightweight method accurately recapitulates protein flexibility, performing competitively with state-of-the-art models. Our study provides fundamental insights into the nature of discrete protein structure representations and introduces a powerful, near-instantaneous method for modeling protein dynamics.
Source code is available in~\href{https://github.com/zj-liu/TokenMD}{https://github.com/IDEA-XL/TokenMD}.
\end{abstract}

\section{Introduction}
\label{introduction}


The convergence of deep learning and vast protein databases has given rise to powerful protein models that can decipher the intricate rules governing protein sequence, structure, and function~\cite{xu2023protst,varadi2024alphafold,xue2022multimodal}. Trained on billions of protein sequences, protein language models (PLMs) such as ESM demonstrate remarkable transfer learning capabilities across downstream tasks~\cite{lin2023evolutionary}. The rapid development of protein structure prediction models, such as AlphaFold, solves the long-standing challenge of predicting static 3D protein structures with remarkable accuracy~\cite{jumper2021highly}. 

While these breakthroughs are powerful, they largely treat sequence and structure as separate domains. In many applications, especially in protein design tasks like binder design~\cite{watson2023novo} and functional site scaffolding~\cite{wang2022scaffolding}, it requires joint understanding and generation of both modalities. 
This highlights the need for multi-modal models that jointly process protein one-dimensional sequences and three-dimensional structures~\cite{wang2024dplm}. A fundamental obstacle in developing such models is how to combine complex, continuous structural data with discrete amino acid tokens in a unified representation suitable for deep learning. To overcome this issue, recent approaches have converged on the concept of \textbf{protein structure tokenization}, discretizing the continuous 3D space into a finite vocabulary using techniques like the Vector Quantized Variational Autoencoder (VQ-VAE)~\cite{hayes2025simulating,van2024fast,gao2025foldtoken}. 
This approach enables modeling the sequence of amino acids and protein structure in a unified language model~\cite{wang2024dplm}.

Despite the promise of this paradigm, several fundamental questions remain unanswered. First, what is the most effective way to integrate the distinct modalities of protein sequence and discrete structure within a single generative framework? While a simple multilayer perceptron (MLP) adaptor is an intuitive starting point, it may not adequately bridge the gap between these different informational streams. Second, the intrinsic properties of the learned structural vocabularies are largely unexplored. Are these tokens distinct and orthogonal, or have the models learned a robust and potentially redundant set of representations? Understanding the ``grammar'' and ``synonymy'' of this structural language is crucial for interpreting and improving these models.


In this work, we investigate these questions by analyzing the properties of the VQ-VAE structural tokens and their application in structure prediction with a GPT-based generative model. We first demonstrate that the method of integrating sequence and structure information is critical, with pre-trained ESM3 sequence embeddings outperforming original ProGen2 sequence embeddings for accurate structure prediction. We then provide direct evidence of semantic redundancy within the structural codebook, showing that distinct tokens often decode to nearly identical structures. 
The semantic redundancy of the codebook, which is a ``flaw'' for next-token prediction, actually can be employed to explore the flexibility of protein structures. This naturally leads us to study a compelling question: \textit{can the discrete representations learned by the VQ-VAE be leveraged for tasks beyond static prediction, offering a new avenue to model protein dynamics?}
By creating a ``synonym dictionary'' based on this redundancy, we introduce a novel ``synonym swap'' strategy. Our results show that this method can generate conformational ensembles whose statistical properties, measured by Root Mean Square Fluctuation (RMSF), are highly correlated with those from traditional MD simulations. This study, therefore, not only sheds light on the nature of discrete structural representations but also establishes a computationally efficient method for generating realistic protein conformational ensembles, opening new possibilities for the study of protein dynamics.
\section{Preliminary and Related Works}
\label{sec:preliminary}

\begin{figure}[htb]
     \centering
     \includegraphics[width=\columnwidth]{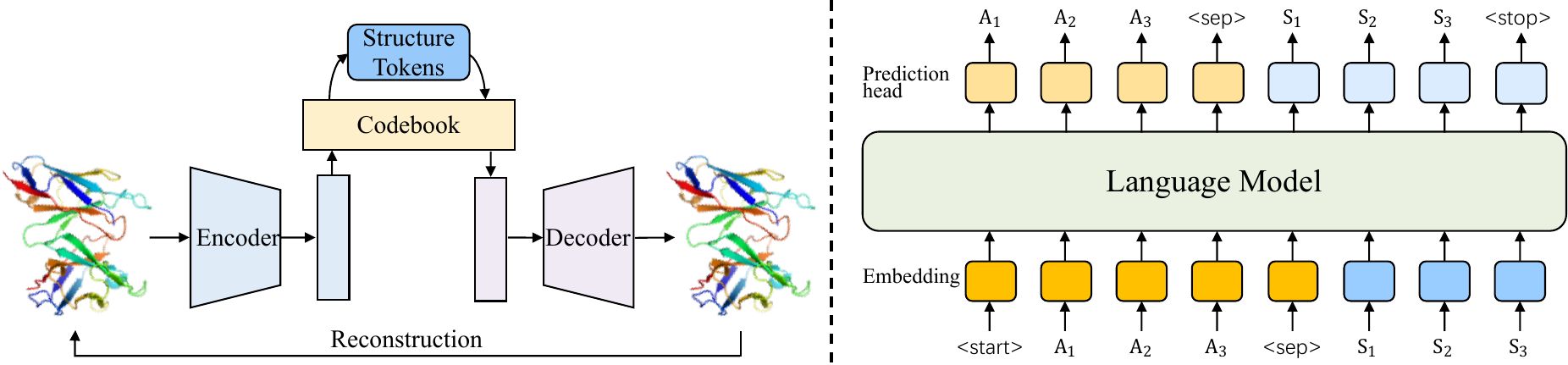}
     \caption{The VQ-VAE (left) discretizes continuous protein structures into a finite set of ``Structure Tokens''. These tokens are then used in an autoregressive language model (right) that predicts a sequence of structural tokens conditioned on the amino acid sequence.}
     \label{fig:overall}
\end{figure}

\subsection{Discrete representation of protein structures}

The discretization of continuous 3D protein structures into a finite set of tokens has emerged as a powerful strategy for applying natural language processing techniques to protein science. Traditional methods usually rely on the domain knowledge about protein structures and discretize protein structures with hard-coded rules~\cite{de2005new,durairaj2020geometricus}.
Recently, a particularly effective approach for this task has been the Vector Quantized Variational Autoencoder (VQ-VAE), which learns a ``vocabulary'' of discrete tokens representing local structural motifs~\cite{van2017neural}. Recent works use the machine-learned vocabulary to show that complex protein folds could be successfully represented as sequences of these learned tokens~\cite{zhang2024balancing,hayes2025simulating}.

As shown in Figure~\ref{fig:overall}, the process of VQ-VAE for protein structure consists of three key components. 
\begin{itemize}
    \item Encoder: A neural network that takes as input the atomic coordinates of the protein and compresses this geometric information into a continuous latent vector.
    \item Codebook: The continuous vector from the encoder is mapped to the nearest vector in a learned codebook via a nearest-neighbor lookup. The index of this codebook vector becomes the discrete structural token.
    \item Decoder: A second neural network that takes a sequence of discrete tokens, retrieves their corresponding vectors from the codebook, and uses this sequence of embeddings to reconstruct the protein structure.
\end{itemize}

It is important to note that the VQ-VAE learns this ``vocabulary'' of discrete tokens purely from the 3D atomic coordinates. Therefore, the structural tokens represent local geometric motifs, independent of the amino acid identities at those positions.
These structural vocabularies can serve as a fundamental component of a unified generative model, including multimodal diffusion language models, masked language models, and autoregressive GPT models~\cite{wang2024dplm,hayes2025simulating,gao2025foldtoken}, enabling sophisticated protein design and analysis.
There are also efforts focused on systematically benchmarking different tokenization schemes and developing improved recipes to guide future research~\cite{yuan2025protein}.

\subsection{Autoregressive sequence-structure language modeling}

With a protein structure tokenizer, we can represent both the protein sequence and its corresponding structure as discrete tokens. A unified generative model can then be developed. This paradigm enables the joint modeling of these two disparate modalities, facilitating tasks such as protein folding and de novo protein design. 

In the context of protein structure prediction, the goal is to model the conditional probability of the structural token sequence ($S_{struct}$) given the amino acid sequence ($S_{seq}$), denoted as $P(S_{struct} | S_{seq})$. An autoregressive model, such as GPT~\cite{radford2019language}, can be used to model $P(S_{struct} | S_{seq})$ by factorizing the probability sequentially:

\begin{align}
\label{eq:1}
    P(S_{struct} | S_{seq}) = \prod_{t=1}^{L} P(s_t | S_{seq}, s_{<t}),
\end{align}

where $s_t$ is the structural token at position $t$, and $s_{<t}$ represents all the preceding structural tokens. At each step of the generation process, the model takes the full amino acid sequence and the sequence of structural tokens predicted so far as input. It then outputs a probability distribution over the entire structural token vocabulary for the next position, $t$. A token is sampled from this distribution, appended to the sequence of predictions, and the process is repeated until the full structure is generated. This generative framework enables the direct prediction of a protein's 3D structure from its primary sequence, forming the basis of the predictive model used in this study.

\subsection{Conformational Ensemble Generation}
After the release of AlphaFold2, many works have focused on the modeling of conformational ensembles other than single-structure prediction~\cite{saldano2022impact,bryant2024structure,xie2024can,brotzakis2025alphafold}. One of the successful explorations is the approach of MSA subsampling~\cite{del2022sampling}, which has been employed to study different protein conformational states ~\cite{casadevall2023alphafold2}. There are also other methods based on AlphaFold2 for sampling multiple protein conformations, such as MSA clustering~\cite{wayment2024predicting} and MSA mutations~\cite{stein2022speach_af}. In addition to exploiting the ability of AlphaFold2, several approaches based on diffusion models are proposed to generate protein ensembles, either in the form of sequence to ensemble~\cite {jing2023eigenfold,jing2024alphafold} or structure to ensemble~\cite{zheng2024predicting,lu2023str2str}. However, these models typically require extensive training on massive datasets to obtain a decent performance. Unlike the methods described above, our ``synonym swap'' approach is (1) training-free, (2) MSA-free, and (3) near-instantaneous. It does not rely on physics simulations, MSA sampling, or training on MD data. Instead, it leverages an intrinsic property, the semantic redundancy, of a pre-trained VQ-VAE codebook to heuristically sample the local conformational space.
\section{The gap between the structure and sequence semantics}
\label{sec:result1}

\subsection{A study of structure tokenization with a GPT-like model for protein structure prediction}

\textbf{Model architecture.} Given the discrete tokenization of the protein structures, it is a natural choice for jointly modeling the protein structure and sequence with a causal language model. In this work, we use the structural VQ-VAE from ESM3 as the structure tokenizer~\cite{hayes2025simulating}.
Following~\citet{liu2023visual}, the network architecture is shown in Figure 1. We choose ProGen2-medium (764M)~\cite{nijkamp2023progen2} as the core autoregressive protein language model.

For the input structure tokens, we use a simple linear layer to align their embeddings with the PLM's embedding space. Specifically, the input structure tokens are first passed through an embedding layer to get a continuous presentation $Z$. A trainable projection matrix $\mathbf{W}_{struct}$ is then applied to convert $Z$ to the same dimensionality as the PLM embedding space. As the model needs to predict the structure tokens, a simple linear layer with weight $\mathbf{W}_{head}$ is added to the PLM as a new prediction head. 

To bridge the semantic gap between the sequence and structure modalities, we investigate two settings for the input sequence embeddings. The first is to use the original \texttt{nn.embedding} layer of ProGen2. Considering the gap between the sequence and structure modalities, the second setting is using the pre-trained sequence embeddings from the ESM3 (1.4B-open) model, which are then followed by a simple linear aligner with a weight matrix $\mathbf{W}_{seq}$ to match the ProGen2 embedding space.

\textbf{Two-stage training.}
To align the structure token embeddings with the PLM embedding, we first keep the PLM weights frozen and train the projection matrix $\mathbf{W}_{struct}$ (and $\mathbf{W}_{seq}$ and in the second setting) and the structure head matrix $\mathbf{W}_{head}$ by maximizing the likelihood of Eq.~\ref{eq:1}.
After embedding alignment, we perform a full fine-tuning to update both the weights of ProGen2 and the linear layers trained in the first stage. 

\textbf{Training data.}
For the first-stage training, we utilize the AFDB SwissProt data~\cite{varadi2024alphafold}. 
For the second stage fine-tuning, we use both AFDB structures and the single-chain structures from PDB. Since the ProGen2 model has a maximal sequence length of 1024, we crop the sequences to a maximal length of 512 to model the sequence and structure tokens together.

\textbf{Structure prediction performance.}
After training, we generate the structure tokens conditioned on the sequence tokens using top-p sampling~\cite{holtzman2019curious} with $p=0.9$ and temperature $T=0.7$. The generated structure tokens are then decoded to obtain the protein structure prediction. This GPT-based architecture bypasses the MSA generation and search steps and can predict the structure of a 200-residue protein in less than 5 seconds, achieving a 100+ fold speedup over the standard AlphaFold2 pipeline. A more important observation, however, is from the structure prediction performance, which is evaluated on three datasets, including CAMEO, CASP14, and CASP15~\cite{haas2018continuous,kryshtafovych2021critical,kryshtafovych2023critical}. Our experiments reveal a stark difference in performance between the two choices of sequence embedding (Table~\ref{tab:tmscore}). The model trained using the original ProGen2 sequence embedding fails to produce globally coherent structures. The model trained using the ESM3 embeddings demonstrates strong predictive performance, comparable to the open 1.4B ESM3~\cite{hayes2025simulating}, as measured by the TM-score~\cite{zhang2004scoring} and the root-mean-square deviation (RMSD)~\cite{kabsch1978discussion}.

\begin{figure}[htb]
     \centering
     \includegraphics[width=0.7\columnwidth]{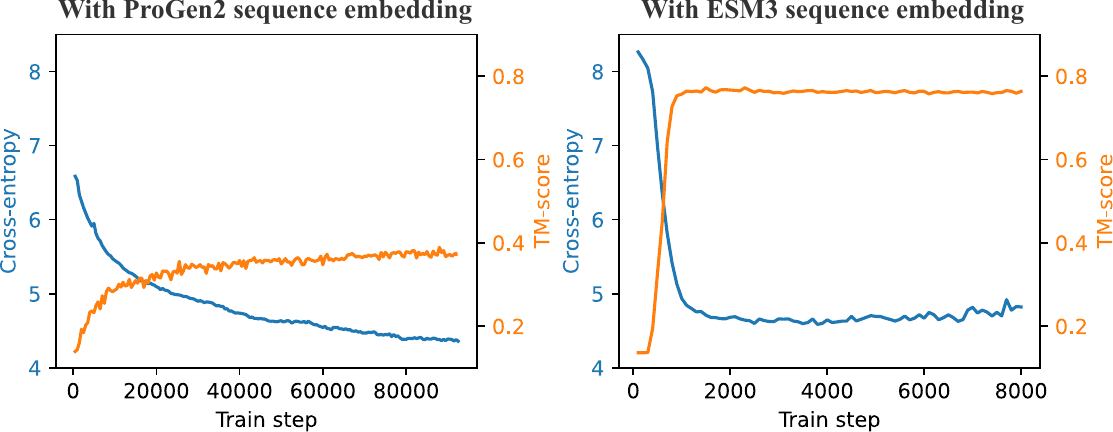}
     \caption{The training curve of the GPT model for protein structure prediction with different sequence embeddings.}
     \label{fig:training}
\end{figure}

By looking into the training curve (Figure~\ref{fig:training}), we find that with the original ProGen2 sequence embedding, while the cross-entropy of the CAMEO test set (i.e., the testing loss) continues to decrease, the structural accuracy of the generated proteins, as measured by TM-score, although still increasing very slowing, stays at a poor performance level with near 90,000 training steps. In contrast, the training curve with the ESM3 sequence embedding shows much faster convergence and a more stable performance plateau. Moreover, the two cases have similar cross-entropy losses. This observation leads us to hypothesize that the training objective itself might be complicated by the nature of the structural vocabulary. Specifically, the size of the ESM3 VQ-VAE codebook is relatively large ($4096$). If distinct tokens can represent structurally similar geometries, the model would be penalized during training for predicting a valid ``structural synonym'' that deviates from the one specific token in the reference structure. This suggests the existence of semantic redundancy within the VQ-VAE's learned vocabulary. 


\begin{table}[t]
\caption{The performance of structure prediction. StructGPT: ProGen2 model with the original ProGen2 sequence embedding; StructGPT w. ESM3\_emb: ProGen2 model with the ESM3 sequence embedding. The results of ESM3 are from~\citet{zhang2024balancing}}
\label{tab:tmscore}
\begin{center}
\resizebox{0.7\linewidth}{!}{
\begin{tabular}{l|ccc|ccc}
  & \multicolumn{3}{c|}{TMScore} &  \multicolumn{3}{c}{RMSD (Å)}  \\
\toprule
Model  & CAMEO & CASP14 & CASP15 & CAMEO & CASP14 & CASP15  \\
\midrule
StructGPT    & 0.523 & 0.329 & 0.383 & 11.87 & 17.19 & 18.99 \\
StructGPT w. ESM3\_emb & 0.784 & 0.580 & 0.639 & 5.43 & 10.24 & 11.74 \\
ESM3-sm-open   & 0.781 & 0.575 & 0.625 & 5.74 & 10.29 & 14.69 \\
\bottomrule
\end{tabular}
}
\end{center}
\end{table}

\subsection{Semantic redundancy in the ESM3 structural codebook}

From the above result, we hypothesize that the ESM3 VQ-VAE codebook contains multiple tokens for similar geometric motifs. We thus turn to directly analyzing the properties of the structural vocabulary to test this hypothesis.
To quantify the similarity between different structural tokens, we perform a direct analysis of the learned codebook. The latent vectors, corresponding to the structure tokens, are extracted from the ESM3 structure VQ-VAE model.

\begin{figure}[h]
     \centering
     \includegraphics[width=\columnwidth]{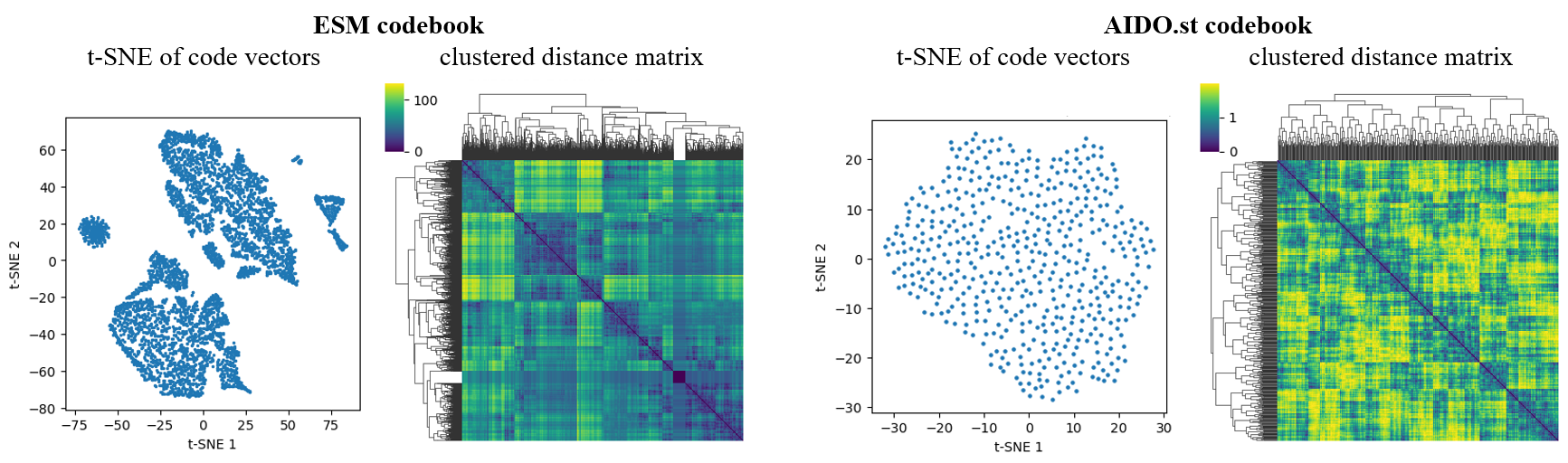}
     \caption{t-SNE visualization and the distance matrix of the code vectors. The ESM3 codebook shows distinct, well-defined clusters, while the t-SNE of the AIDO.st codebook~\cite{zhang2024balancing} vectors are uniformly distributed in the 2D space.}
     \label{fig:tsne}
\end{figure}

First, to visually inspect the relationships within the tokens, we use the t-SNE (t-Distributed Stochastic Neighbor Embedding) visualization~\cite{maaten2008visualizing}, which projects the high-dimensional token embeddings into a 2D space. As shown in Figure~\ref{fig:tsne}, the t-SNE projection illustrates the existence of numerous dense clusters, indicating there are grouped tokens representing similar structural motifs in the embedding space.

Second, to establish a quantitative measure of similarity, we calculate the pairwise Euclidean distance between all token latent vectors. From the clustered heatmap, it is easy to see that the ESM3 structure tokens have strong, well-separated clusters. In contrast, the AIDO.st VQ-VAE codebook~\cite{zhang2024balancing}, which has a smaller number of codes ($m=512$), shows a more diffuse and less defined clustering pattern.

\begin{wraptable}{hbt}{0.35\textwidth}
    \caption{The average TM-score and RMSD of the structures decoded from the perturbed structure tokens.}
    \label{tab:perturb}
    \centering
    \resizebox{\linewidth}{!}{
    \begin{tabular}{c|c|c}
     & \multicolumn{1}{c|}{TM-score} & \multicolumn{1}{c}{RMSD (Å)} \\
    \midrule
     CAMEO & 0.933 & 1.744 \\ 
     \hline
     CASP14 & 0.938 & 1.969 \\ 
     \hline
     CASP15 & 0.849 & 4.136 \\ 
    \bottomrule
    \end{tabular}
    }
\end{wraptable}

Based on this, we construct a ``synonym dictionary'' by defining any two tokens as synonyms if the Euclidean distance between their latent vectors is below a threshold. Specifically, given a codebook $\mathcal{V} = \{ \mathbf{v}_k\}_{k=1}^{m} $ of $m$ different codes, the ``synonym dictionary'' of code $\mathbf{v}_k$ is defined by $\mathcal{S}_k = \{i\}_{\Vert \mathbf{v}_i - \mathbf{v}_k \Vert_2 < \tau }$, where $\tau$ is a threshold hyperparameter that can balance structural preservation and diversity. Based on visual inspection of the distance distribution, we currently set $\tau=10$. 
The dictionary $\mathcal{S}_k$ serves as the basis for our subsequent perturbation studies.
Given a structure, we first encode it into a sequence of structure tokens. Each token is then replaced by a random token in its ``synonym dictionary'' and the perturbed structural sequence is decoded into 3D structures. The resulting 3D structures are very similar to the original structure. The RMSD between the perturbed and original structures is consistently low, often less than 2.0 Å, confirming their structural equivalence. This indicates the semantic redundancy of the ESM3 codebook, which explains why a GPT model is statistically uncertain about which specific token to predict next, yet still can point towards the correct local geometry. 

\section{Exploiting redundancy to generate dynamic conformational ensembles}
\label{sec:result2}


Our finding in the previous section demonstrates that the structural vocabulary of ESM3 is not a minimal set of building blocks but a robust, flexible, and highly redundant language. The semantic redundancy of the codebook, which is a flaw for next-token prediction, is actually a feature that reflects the inherent flexibility of protein structures. Together with the structure decoder, the subtle structural variations in the ``synonymous'' tokens may reflect the natural, low-energy fluctuations a protein experiences in its native state.
If this hypothesis is true, perturbing a ground truth structure by swapping its tokens with synonyms could provide a computationally inexpensive method to generate a realistic conformational ensemble, offering a rapid alternative to Molecular Dynamics (MD) simulations for studying protein flexibility~\cite{abramson2024accurate}.

\subsection{Method}
\begin{figure}[htb]
     \centering
     \includegraphics[width=0.7\columnwidth]{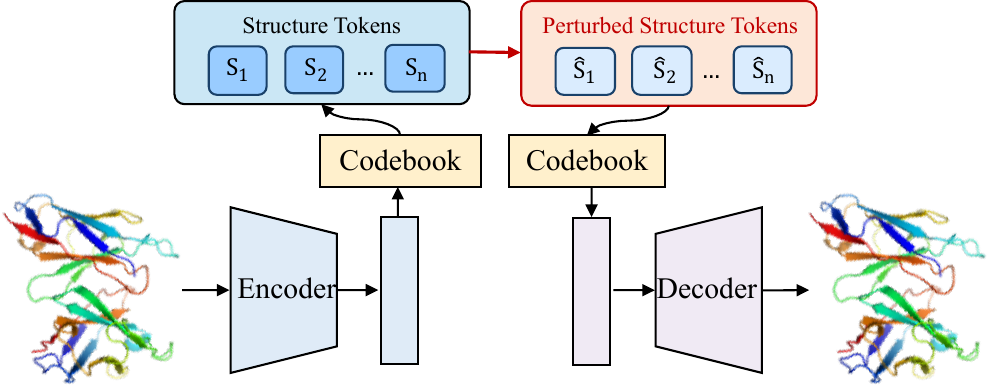}
     \caption{Perturbation of the structure tokens for exploring the conformational ensemble space.}
     \label{fig:perturb}
\end{figure}

To test our hypothesis, we employ a simple ``synonym swap'' strategy. As shown in Figure~\ref{fig:perturb}, we first encode a given experimental structure of a target protein into a sequence of structure tokens. 
A new sequence of structure tokens is then generated by randomly replacing each token $k$ from the original sequence with one of its synonyms, as defined in the synonym dictionary $\mathcal{S}_k$. 
The perturbed token sequence is then decoded back into a 3D protein structure. 

To validate our approach, we use the 82 test proteins~\cite{jing2024alphafold} from the ATLAS database~\cite{vander2024atlas}. Following the work in~\citet{jing2024alphafold}, we generate 250 perturbed structures for each target as the conformational ensembles.

\subsection{Experiment}

\begin{wraptable}{ht}{0.55\textwidth}
\vspace{-0.4cm}
    \caption{The median of results on the 82 test targets in ATLAS; the results of MDGen and AlphaFlow are from~\citet{jing2024alphafold,jing2024generative}}
    \label{tab:rmsf}
    \centering
    \resizebox{\linewidth}{!}{
    \begin{tabular}{c|ccc}
    \toprule
     & \hspace{2mm}Ours\hspace{2mm} & MDGen& AlphaFlow \\
    \midrule
     Pairwise RMSD $r$ $\uparrow$ & 0.38 & 0.48 & 0.48 \\ 
     \hline
     Per target RMSF $r$ $\uparrow$ & 0.84 & 0.71 & 0.85\\ 
     \hline
     Global RMSF $r$ $\uparrow$& 0.41 & 0.50 & 0.50\\ 
     \hline
     MD PCA $\mathcal{W}_2$ dist. $\downarrow$
 & 1.83 & 1.89 & 1.52 \\ 
     \hline
     Joint PCA $\mathcal{W}_2$ dist. $\downarrow$ & 2.54 & - & 2.25 \\ 
    \bottomrule
    \end{tabular}
    }
    \vspace{-0.7cm} 
\end{wraptable}

To evaluate the quality of our generated ensembles, we compare our ``synonym swap'' method to two state-of-the-art models, MDGen and AlphaFlow~\cite{jing2024alphafold,jing2024generative}, using the ground-truth trajectories from Molecular Dynamics (MD) simulations. 
To compare two ensembles, we use a suite of metrics designed to assess both protein flexibility and the conformational distribution. The results, summarized in Table~\ref{tab:rmsf}, demonstrate that our computationally efficient method is highly competitive, particularly in capturing protein-specific flexibility.

A key measure of an ensemble's utility is its ability to reproduce the flexibility profile of individual proteins. The root mean square fluctuation (RMSF) quantifies the fluctuation of individual residues.
The median Pearson correlation ($r$) between the generated and MD RMSF for each protein is 0.84. This result is highly competitive with the top-performing method, AlphaFlow (0.85), and outperforms MDGen (0.71), confirming that token perturbation effectively captures the unique dynamic fingerprint of individual proteins. Note that MDGen and AlphaFlow are both trained on the ATALS MD data, while our ``synonym swap'' method is totally \textbf{training-free}.

To assess how well the ensembles capture the dominant modes of motion, we use the 2-Wasserstein distance ($\mathcal{W}_2$) between the generated and MD distributions after projecting them onto the principal components (PCs) of the positional distribution. A lower distance indicates a better match. Our method achieved an MD PCA $\mathcal{W}_2$ distance of 1.83, slightly better than MDGen (1.89) and competitive with AlphaFlow (1.52), indicating that the generated conformations occupy the same principal dynamic spaces as those explored by the MD simulation. 

\begin{wrapfigure}{ht}{0.6\textwidth}
     \centering
     \includegraphics[width=0.6\columnwidth]{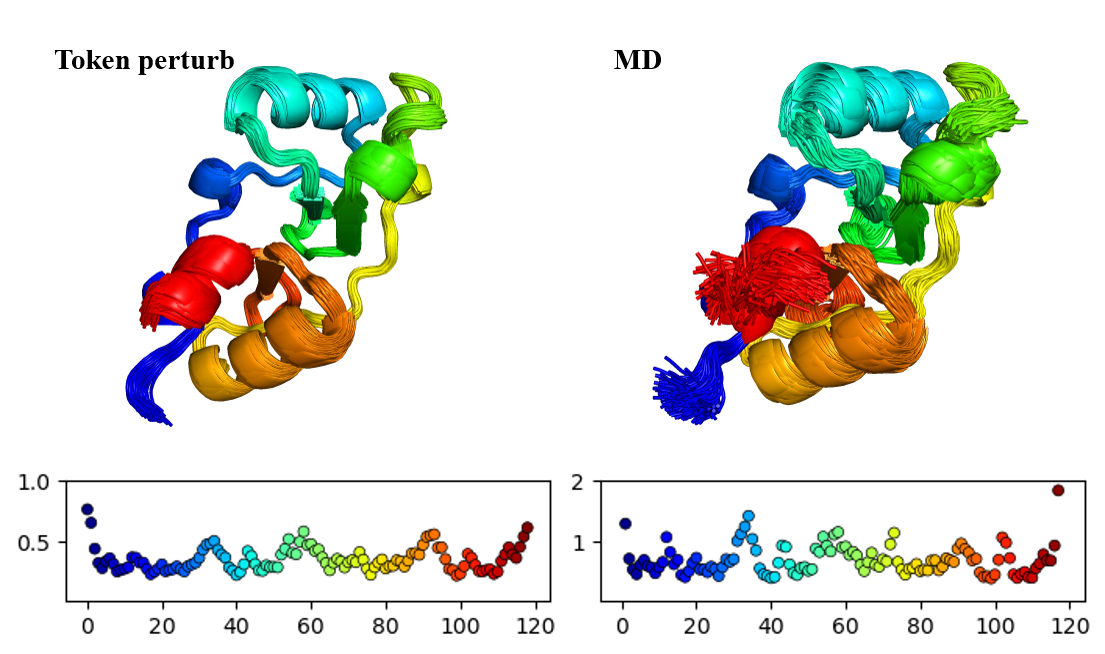}
     \caption{Protein ensembles for \texttt{6uof\_A} generated by token perturbation and MD, and the C$\alpha$ RMSFs indexed by the residue id (Pearson $r=0.81$).}
     \label{fig:rmsf}

\end{wrapfigure}

While our method can accurately capture per-protein flexibility and conformational distribution, other metrics that assess the global properties of the conformational space show the limitation of our current simple approach. The correlation of the Pairwise RMSD matrices and the Global RMSF correlation is lower for our method compared to the other two benchmarked methods. This suggests that while individual flexibility profiles are accurate, capturing the absolute scale of motion across a diverse dataset is more challenging (Figure~\ref{fig:rmsf}). 
Our ``synonym swap'' strategy is a heuristic, training-free approach. It is not designed to replace physics-based simulations or capture large-scale, cooperative motions. Instead, its primary advantage is its computational efficiency—offering a near-instantaneous method to generate a plausible conformational ensemble that reflects the local flexibility and low-energy fluctuations inherent in a protein's native state.
The local perturbations created by ``synonym swapping'' may not be sufficient to capture the large-scale, cooperative motions that dictate global structural changes. This provides a direction for future work, such as exploring perturbations of token sequences rather than individual tokens to model more complex dynamics or combining with other techniques like MSA subsampling~\cite{del2022sampling}.

\section{Conclusion}

In this paper, we investigate the intrinsic properties of these discrete structural representations and explore how they can be leveraged for tasks beyond static prediction. We first develop a GPT-based single sequence structure predictor and show that a GPT model using pre-trained ESM3 sequence embeddings significantly outperforms one using simple token embeddings. From this observation, we then demonstrate that the structural codebook contains considerable semantic redundancy, where distinct tokens decode to nearly identical local structures. Finally, we harness this redundancy by developing a ``synonym swa'' strategy, showing it can generate conformational ensembles whose dynamic properties are highly correlated with those from computationally expensive Molecular Dynamics (MD) simulations~\cite{dror2012biomolecular,miller2021moving}. Our findings provide a deeper understanding of discrete structural representations and offer a novel, efficient method for modeling protein dynamics.

\section*{Acknowledgement}
This work was supported by Shenzhen Hetao Shenzhen-Hong Kong Science and Technology Innovation Cooperation Zone under Grant No. HTHZQSWS-KCCYB-2023052.

\bibliographystyle{neurips_2025}
\bibliography{ref}




\end{document}